\documentclass[10pt,twocolumn,letterpaper]{article}

\usepackage{cvpr}              % To produce the CAMERA-READY version
\usepackage{times}
\usepackage{epsfig}
\usepackage{graphicx}
\usepackage{amsmath}
\usepackage{amssymb}
\usepackage{booktabs}
\usepackage{pifont}
\usepackage{soul}
\usepackage{stfloats}
\usepackage[dvipsnames]{xcolor}

\definecolor{cvprblue}{rgb}{0.21,0.49,0.74}
\usepackage[pagebackref,breaklinks,colorlinks,citecolor=cvprblue]{hyperref}

\def\paperID{8999} % *** Enter the Paper ID here
\def\confName{CVPR}
\def\confYear{2024}

\title{SurroundSDF: Implicit 3D Scene Understanding Based on Signed Distance Field}

\author{
Lizhe Liu$^{1*}$\hspace{3mm}
Bohua Wang$^{2* \S}$ \hspace{3mm}
Hongwei Xie$^{1*}$\hspace{3mm}
Daqi Liu$^{1}$\hspace{3mm} 
Li Liu$^{1}$\hspace{3mm}
Zhiqiang Tian$^{2}$
\\
Kuiyuan Yang$^{1}$\hspace{3mm}
Bing Wang$^{1\dagger}$
\\ 
${}^{1}$Xiaomi EV \hspace{6mm}
${}^{2}$School of Software Engineering, Xi'an Jiaotong University
\\
}
\begin{document}
\maketitle
\begin{abstract}
Vision-centric 3D environment understanding is both vital and challenging for autonomous driving systems. Recently, object-free methods have attracted considerable attention. Such methods perceive the world by predicting the semantics of discrete voxel grids but fail to construct continuous and accurate obstacle surfaces. To this end, in this paper, we propose \textbf{SurroundSDF} to implicitly predict the signed distance field (SDF) and semantic field for the continuous perception from surround images. Specifically, we introduce a query-based approach and utilize SDF constrained by the Eikonal formulation to accurately describe the surfaces of obstacles. Furthermore, considering the absence of precise SDF ground truth, we propose a novel weakly supervised paradigm for SDF, referred to as the \textbf{Sandwich Eikonal} formulation, which emphasizes applying correct and dense constraints on both sides of the surface, thereby enhancing the perceptual accuracy of the surface.
Experiments suggest that our method achieves SOTA for both occupancy prediction and 3D scene reconstruction tasks on the nuScenes dataset.
\end{abstract}    
\section{Introduction}
\label{sec:intro}

\def\thefootnote{*}\footnotetext{
Equal Contribution.
}
\def\thefootnote{$\S$}\footnotetext{
Work done during an internship at Xiaomi EV.
}
\def\thefootnote{$\dagger$}\footnotetext{
Corresponding Author.
}

With the recent advancement of 3D object detection algorithms \cite{huang2021bevdet, li2023bevdepth, li2022bevformer, liu2022petr, liu2023bevfusion, Huang_Liu_Zhang_Zhang_Xu_Wang_Liu_2022}, vision-centric autonomous driving system become more practicable. Nevertheless, the persisting challenges related to the long-tail problem and the coarse depiction of the 3D scene underscore its insufficiency. Consequently, a deeper comprehension of 3D geometry and semantics is needed for safety and reliability. This paper delves into a novel vision-centric paradigm of dense and continuous 3D scene understanding.

Current approaches of dense 3D prediction can be classified into two categories, \textit{3D reconstruction} and \textit{3D perception}. Specifically, \textit{3D reconstruction} algorithms \cite{guizilini2019semantically, schmied2023r3d3} generate dense point clouds enriched with semantic information by projecting depth maps and 2D semantic maps into 3D space. \textit{3D perception} methods \cite{cao2022monoscene, huang2023tri, tian2023occ3d, wei2023surroundocc, zhang2023occformer, li2023voxformer, Miao_Liu_Chen_Gong_Xu_Hu_Zhou} predict the occupied status and the semantic of 3D voxel grids. However, either method proves to be redundant or can only predict coarse-grained discrete grids.

\begin{figure}
\begin{center}
\includegraphics[scale=0.34]{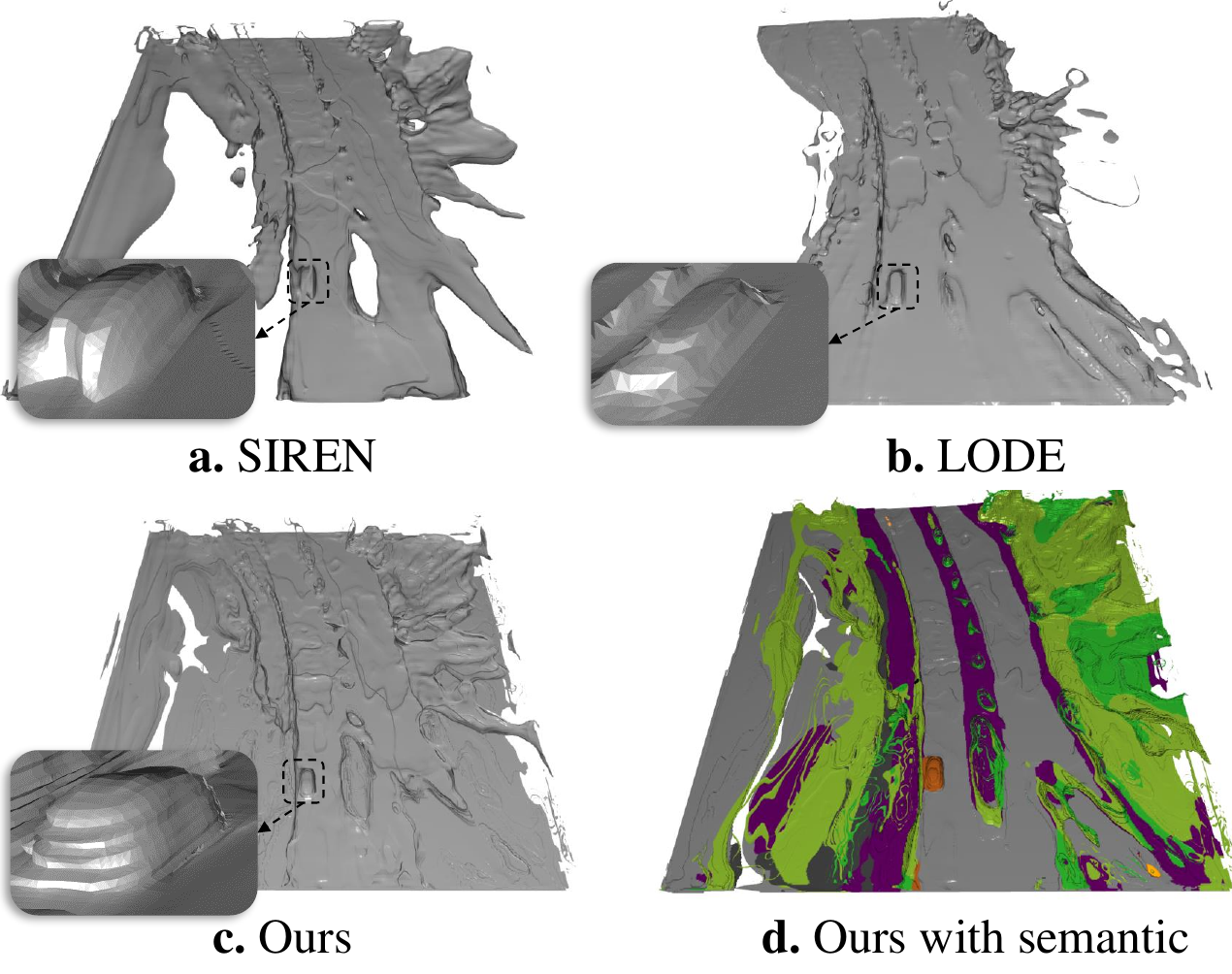}
\end{center}
\vspace{-6mm}
\caption{The scene perspective results from surround images input. \textbf{a, b.} The result of SIREN and LODE supervision with camera-only input (the original methods are point cloud-based). \textbf{c.} Our result. \textbf{d.} Our result with semantics.}
\label{Fig.fig1}
\vspace{-7mm}
\end{figure}

Instead, we develop a vision-centric framework to describe 3D scenes leveraging neural implicit Signed Distance Function (SDF) representation, which we refer to as \textbf{SurroundSDF}. Generally, we aim to: (1) Continuously describe the 3D scene by reconstructing smooth surfaces; (2) Explore the difficulty associated with the utilization of SDF representation and pose an appropriate training strategy; (3) Exploit the strong representation to kill two birds with one stone, addressing 3D semantic segmentation and continuous 3D geometry reconstruction within one framework.

In pursuit of the first and the third objectives, we employ SDF to represent the canonical 3D scene in terms of distance from the surface. Furthermore, we construct an implicit field that encompasses both semantics and geometry by exploiting this strong representation. However, it's still nontrivial to accomplish the SDF modeling, where complete surface and normal vectors, which are crucial supervisions, cannot be accurately computed. To this end, SIREN \cite{sitzmann2020implicit} proposed a weakly supervised implicit perception method, which accomplishes 3D mesh completion from point clouds. While LODE \cite{LiZSZYZZ23} proposed to utilize occupancy ground truth (GT) to supervise SDF. However, these methods rely on the input of point clouds, and only alleviate this difficulty by mimicking the surface distance but are either sparse or inaccurate. 

To overcome the aforementioned challenges, we propose the \textbf{Sandwich Eikonal} formulation, a novel weak supervision paradigm for SDF modeling. Figure~\ref{Fig.fig1} gives some reconstruction results and shows the benefit of our method. This method emphasizes applying correct and dense constraints on both sides of the surface to enhance the geometric accuracy and continuity of the surface. Moreover, we revisit current pipelines for training the perception branch and design a novel loss that enhances the integration of geometry and semantics thereby reducing inconsistencies. Our contributions can be summarized as follows:

\begin{itemize}
\item We propose a vision-centric implicit semantic SDF perception method, achieving accurate and continuous 3D perception. To the best of our knowledge, we are the first to utilize SDF for surround-view 3D perception.

\item We introduce the Sandwich Eikonal formulation, a novel weak supervision paradigm for SDF.

\item We demonstrate how to employ this representation for the reconstruction of continuous 3D geometry with precise semantic information. 

\item We achieve state-of-the-art results on 3D dense semantics perception tasks. Comprehensive experiments on the NuScenes dataset \cite{caesar2020nuscenes} provide extensive validation of our approach.
\end{itemize}
\section{Related Work}
\label{sec:related}

\paragraph{Occupancy Prediction}
Recently, researches \cite{Roldao_de_Charette_Verroust-Blondet_2020, Roldao_de_Charette_Verroust-Blondet_2020, cao2022monoscene, huang2023tri, tian2023occ3d, wei2023surroundocc, zhang2023occformer, li2023voxformer, Miao_Liu_Chen_Gong_Xu_Hu_Zhou} on occupancy prediction have demonstrated advantages in 3D scene understanding. 
Compared to the traditional object detection paradigm \cite{detr3d, carion2020end, liu2023petrv2, xie2022m, li2023bevstereo, huang2021bevdet, li2023bevdepth, li2022bevformer, liu2022petr, liu2023bevfusion, Huang_Liu_Zhang_Zhang_Xu_Wang_Liu_2022}, occupancy perception has the following advantages. First, it can express dense 3D geometry. Second, it can accurately provide spatial locations for objects beyond predefined categories. Third, it can describe the shapes of irregular obstacles. 

Based on these advantages, occupancy prediction tasks have attracted significant research. These methods predict the occupancy status in the region of interest around the ego vehicle from point clouds or images. Specifically, the space is first divided into voxel grids at a specific resolution. Then the occupancy status and semantics of each grid are estimated. SurroundOcc \cite{wei2023surroundocc} proposed a surround-view 3D occupancy perception method that utilizes spatial 2D-3D attention to lift image features into 3D space. In addition, to realize the dense occupancy prediction, SurroundOcc designed a pipeline to convert the point cloud to dense occupancy ground truth. VoxFormer \cite{li2023voxformer} employed an MAE-like \cite{he2022masked} approach to achieve camera-based semantic occupancy prediction. FB-OCC \cite{li2023fb, li2023fb} proposed a novel forward-backward projection method to compensate for the insufficient BEV feature density of the forward projection method and the large number of mismatches in 2D and 3D space caused by the backward projection. Despite the impressive results, they are limited by the specific resolution of the occupancy annotations, which limit the continuous perception of the scene.
\vspace{-4mm}

\begin{figure*}[ht]
\begin{center}
\includegraphics[scale=0.33]{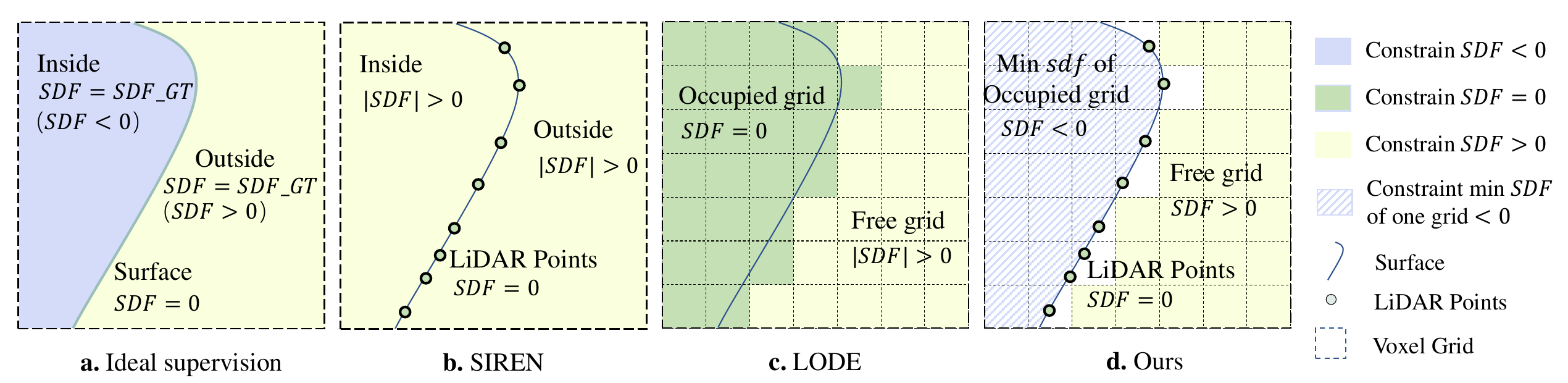}
\end{center}
\vspace{-6mm}
   \caption{\textbf{a.} Ideal supervision for SDF where the SDF GT is provided. \textbf{b.} SIREN supervision which uses LiDAR points to supervise the surface. \textbf{c.} LODE supervision based on occupancy GT. \textbf{d.} Our supervision paradigm which combines the LiDAR GT and occupancy GT, it is closer to the ideal supervision.}
\label{Fig.compare}
\vspace{-5mm}
\end{figure*}

%-------------------------------------------------------------------------
\paragraph{Implicit Scene Perception}
Scene reconstruction refers to the task of predicting the 3D geometry structures from some incomplete representations, e.g., images, point clouds, and voxel grids. Implicit neural scene reconstruction methods \cite{sitzmann2020implicit, LiZSZYZZ23, yuan2022monocular, cao2023scenerf, agro2023implicit, park2019deepsdf, mildenhall2021nerf} have demonstrated advantages in accuracy and substantial potential. Generally, they train a neural network to predict a continuous field. Thus, it is possible to query occupancy information for any point within the 3D space. DeepSDF \cite{park2019deepsdf} utilizes the SDF to achieve the implicit reconstruction of 3D objects at the instance level. NeRFs \cite{mildenhall2021nerf} and its numerous variants \cite{xie2022s, cao2023scenerf, Guo_Deng_Li_Bai_Shi_Wang_Ding_Wang_Li_2023} achieve implicit 3D scene reconstruction and novel view synthesis from multi-view images. NeuS \cite{wang2021neus} and VolSDF \cite{yariv2021volume} introduce the signed distance function into neural rendering, enabling more precise object surface reconstruction.  SIREN \cite{sitzmann2020implicit} utilizes the Eikonal equation to achieve semi-supervised SDF reconstruction for point clouds. However, these methods lack generalization to novel scenes. To achieve online implicit scene perception in driving scenes, LODE \cite{LiZSZYZZ23} proposes the Locally Conditioned Eikonal formulation and introduces a dense occupancy ground truth for supervision, which significantly improves the effect of SDF reconstruction in the driving scenes. However, it is limited by the sparsity of input LiDAR points or inaccurate ground truth. To solve these problems, we propose a camera-only implicit scene understanding method, which aims at continuous and accurate surface perception.

\vspace{-1mm}
\section{Formulation}
\vspace{-1mm}
This section analyzes the SDF supervision paradigm based on the Eikonal formulation with its variations and introduces our \textbf{Sandwich Eikonal} supervision approach. 
\vspace{-1mm}
\subsection{Eikonal-based SDF Constraints}
\vspace{-1mm}
Given a coordinate $\textbf{x}$ in the 3D scene, our goal is to construct a function $\phi$ such that \(\phi(\textbf{x})\) provides the SDF value at point $\textbf{x}$. The Eikonal-based optimization objectives \cite{LiZSZYZZ23} is
\vspace{-2mm}
\begin{equation}
\scalebox{0.9}{$
    \mathcal{O}_{E} = \int_{\Omega_0} O_0 d\textbf{x} + \int_{\Omega_1} O_1 d\textbf{x} + \int_{\Omega_2} O_2 d\textbf{x},
    $}
\label{Eq.eikonal1_objective} 
\end{equation}
and
\begin{equation}
    \scalebox{0.9}{$
    \begin{cases}
      O_0=\left ||\nabla _\textbf{x}\phi(\textbf{x}) |-1  \right |,& \textbf{x}\in \Omega_0 \\
      O_1=\left |\nabla _\textbf{x}\phi(\textbf{x}) -\textbf{n}(\textbf{x})  \right |,& \textbf{x}\in \Omega_1\ \\
      O_2=\left |\phi(\textbf{x}) -SDF(\textbf{x})  \right |,& \textbf{x}\in \Omega_2
    \end{cases},
    $}
\label{Eq.eikonal1} 
\end{equation}

where $\Omega_0$ is the set of the whole 3D space of interest, $\Omega_1$ is the set of points on the surface, $\Omega_2$ is the set of points with SDF GT annotations, $\nabla _\textbf{x}\phi(\textbf{x})$ is the gradient on point $\textbf{x}$, $\textbf{n}(\textbf{x})$ is the normal, and $SDF(\textbf{x})$ is the real SDF value at point $\textbf{x}$.
One major obstacle to this optimization objective lies in the design of $O_2$, which provides direct supervision for SDF as shown in Figure \ref{Fig.compare}.a. This objective requires the annotation of SDF values for each anchor point in the scene, which is monumental and challenging. 

In the absence of precise SDF annotations, SIREN \cite{sitzmann2020implicit} and LODE \cite{LiZSZYZZ23} have implemented weakly supervised paradigm through the following variations of $O_2$,
\vspace{-1mm}
\begin{equation}
    \scalebox{0.9}{$
    \begin{cases}
    O^\prime_{2-0}=|\phi(\textbf{x})|, &\textbf{x}\in  \Omega_1 \\
    O^\prime_{2-1}=\psi(|\phi(\textbf{x})|), &\textbf{x}\in  \Omega_3,
    \end{cases}
    $}
\vspace{-2mm}
\label{Eq.eikonal2} 
\end{equation}
where $\psi(\cdot)$ is a monotonically decreasing function and $\Omega_3\subseteq \Omega_0/\Omega_1$ is the whole space except the surface.

The objective $O^\prime_{2-0}$ aims to constrain the SDF values on the surface to $0$. Considering the absence of precise supervision, $O^\prime_{2-1}$ adopts a fuzzy constraint that pushes the absolute SDF value away from zero in $\Omega_3$. While the constraint in $\Omega_3$ is loose, it is adequate to capture the geometry information of the surface.

 In practical implementation, SIREN \cite{sitzmann2020implicit} samples on-surface LiDAR points in $\Omega_1$  and considers the space excluding the LiDAR points as $\Omega_3$, as is shown in Figure \ref{Fig.compare}.b. When the LiDAR points are dense enough, this strategy can effectively achieve good constraint effects \cite{sitzmann2020implicit}. However, LODE \cite{LiZSZYZZ23} has demonstrated that the sparsity of LiDAR points in the driving scenes results in discontinuities of the estimated mesh. To achieve denser supervision, they use the dense occupancy GT as supervision. This approach treats both the surface and interior of the obstacles as the $\Omega_1$. Specifically, they sample the $\Omega_1$ points as the center of ``occupied" grids, and the $\Omega_3$ points as the center of ``free" grids, as is shown in \ref{Fig.compare}.c. However, this paradigm remains imprecise, as considering the center of the occupied grid to be the surface of objects would introduce errors.

\begin{figure*}[ht]
\begin{center}
\includegraphics[scale=0.75]{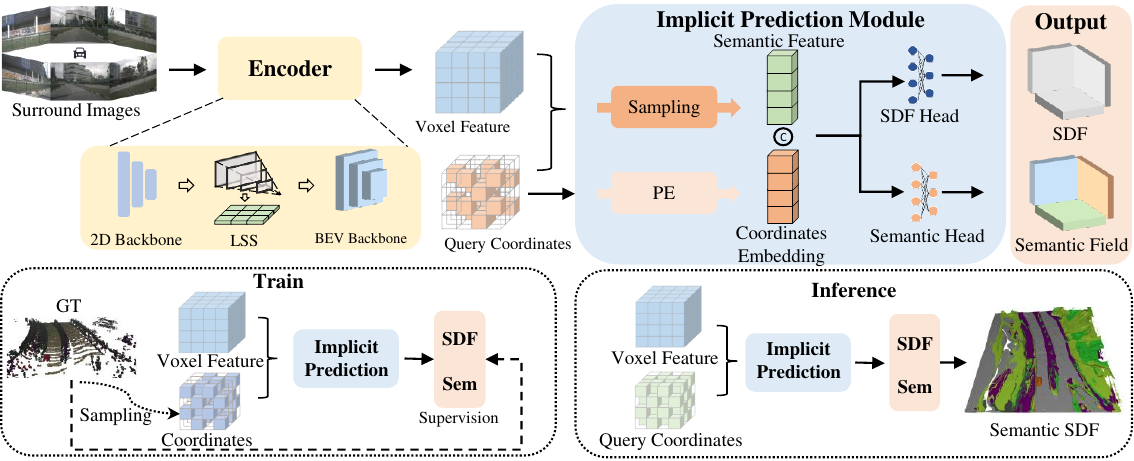}
\end{center}
\vspace{-6mm}
   \caption{The architecture of our SurroundSDF. Given the surround images as input, an encoder composed of a 2D backbone, LSS module, and BEV backbone, is employed to extract voxel features. We adopt a query-based approach to sample features from the voxel features. Specifically, first, a set of query coordinates in the region of interest is selected. Subsequently, using trilinear interpolation, semantic features are queried from the voxel features. Finally, after concatenation with the positional embeddings from the query coordinates, the features pass through the SDF head and semantic head respectively, yielding SDF and semantic fields. For training, the query coordinates are sampled according to the GT, and the SDF and semantic field are supervised by the losses introduced in Section \ref{section:Loss}. In the inference phase, based on appropriate sampling and post-processing, continuous and accurate scene perception results are obtained (see Section \ref{Sec.inference}).}
\vspace{-5mm}
\label{Fig.pipeline}
\end{figure*}

\subsection{Sandwich Eikonal Formulation}
To address the above concerns, we require a supervision paradigm that has the following characteristics: 1) sampling points as the supervision on the surface must strictly adhere to the surface, 2) for regions beyond the surface, refrain from supervision in uncertain areas to ensure precision, and 3) dense supervision signals. Considering the above factors, we introduce our variants of Eikonal formulation as follows:
\vspace{-2mm}
\begin{equation}
    \scalebox{0.9}{$
    \begin{cases}
      O_{2-0}=\phi(\textbf{x}), & \textbf{x}\in \Omega_1 \\
      O_{2-1}=\psi(-min(\phi(\textbf{x}))), & \textbf{x}\in \Omega^i_{occ} \\
      O_{2-2}=\psi(\phi(\textbf{x})), & \textbf{x}\in \Omega^j_{free} ,
      
    \end{cases}
    $}
\vspace{-1mm}
\label{Eq.eikonal4} 
\end{equation}
where $\Omega^i_{occ}$ is the  $i$-th grid of occupied grids and  $\Omega^j_{free}$ is the  $j$-th grid of free grids.

To this end, we incorporate the LiDAR points and the occupancy GT into the SDF supervision. As is shown in Figure \ref{Fig.compare}.d, for precise sampling points on the surface, we follow SIREN and constrain the SDF value of LiDAR points to be $0$. 
To apply accurate and dense supervision to regions beyond the surface, we introduce the occupancy GT into the supervision, as is shown in optimization objectives $O_{2-1}$ and $O_{2-2}$. According to the generation process of occupancy GT \cite{tong2023scene, tian2023occ3d}, if a grid is labeled as ``occupied", it implies that at least part of the grid area falls inside the object. Therefore, as the objective $O_{2-1}$ shows, we constrain the minimum SDF value of each occupied grid to be less than zero and push it to the negative range. 
On the other hand, if a grid is labeled as ``free", the entire grid is outside the object and we push the SDF value of free grids to the positive range, as is shown in objective $O_{2-2}$. We employ a multi-frame point cloud fusion strategy to densify the sampling for surface supervision, as described in Section \ref{section:Loss}. Compared to previous supervision paradigms, our formulation emphasizes applying appropriate and dense supervision on both sides of the surface to enhance the geometric constraint based on the Eikonal formulation. Therefore, we name our supervision strategy the ``\textbf{Sandwich Eikonal}" formulation.
\section{Method}

Given multi-camera images $\textbf{I}=\{I_1, I_2, ...  ,I_n\}$ , we aim to predict the 3D SDF $\phi(\textbf{x}):\mathbb{R}^3\Rightarrow\mathbb{R}$ and semantic field $S(\textbf{x}):\mathbb{R}^3\Rightarrow\mathbb{R}^{s}$, where $s$ indicates the number of classes. The following will introduce our architecture and detailed constraints based on our Sandwich Eikonal formulation.

\subsection{Architecture}
\label{Sec.arch}
The overall architecture is shown in Figure \ref{Fig.pipeline}. First, a CNN-based backbone is used to obtain the image features. Then, the LSS \cite{philion2020lift} is used to project the image features to BEV space, and a BEV encoder is adopted to obtain the voxel features $V$ of dimension $h \times w \times d \times c$ (c indicates the number of channels). Further, for each point \textbf{x} in $\Omega_0$, the corresponding features can be obtained by querying $V$ with the trilinear sampling method. Therefore, $\phi(\textbf{x})$ and $S(\textbf{x})$ can be expressed as:

\vspace{-4mm}
\begin{equation}
    \begin{split}
    \phi(\textbf{x}) = H_{sdf}(C(Q(V,\textbf{x}), PE(\textbf{x})), \theta), \textbf{x}\in \Omega_0, \\
    S(\textbf{x}) = H_{sem}(C(Q(V,\textbf{x}), PE(\textbf{x})), \theta), \textbf{x}\in \Omega_0,
    \end{split}
\label{cacl_sdf_sem} 
\vspace{-4mm}
\end{equation}

 where $\textbf{x}\in\mathbb{R}^3$, $C$ and $Q$ represent the concatenation and query process, respectively. $H_{sdf}$ and $H_{sem}$ indicate the SDF head and semantic head, implemented with three-layer MLPs with sine activation functions. $\theta$ represents the learnable parameters. $PE$ represents the positional encoding, which aims to capture high-frequency information. Specifically, for a coordinate $\textbf{x}\in\mathbb{R}^3$, $PE$ encodes each of its dimensions into a 2n-dimensional vector, it can be expressed as:
 \vspace{-3mm}
\begin{equation}
    PE(\textbf{x}) = (\zeta(x), \zeta(y), \zeta(z)) \in\mathbb{R}^{6n},
\label{pe} 
\vspace{-1mm}
\end{equation}
 where $x,y,z$ denote the coordinates, $\zeta(\cdot)$ represents an encoding scheme using trigonometric function mapping:
\vspace{-1mm}
\begin{equation}
    \zeta(x) = (sin(2^0\pi x),cos(2^0\pi x), ..., cos(2^{n-1}\pi x)).
\label{pe_2} 
 \vspace{-1mm}
\end{equation}

In training, we design a joint supervision method that utilizes both coarse-grained voxel and fine-grained point cloud supervision based on our Sandwich Eikonal formulation.
In inference, we query the voxel features to estimate the SDF values and semantic logits of each query point. 

\subsection{Supervision}
\label{section:Loss}
This section will elaborate on deriving the SDF loss from our Sandwich Eikonal formulation. Then, we will introduce the semantic loss and our joint supervision strategy.

\vspace{-3mm}
\paragraph{SDF Supervision} 
Following objectives $O_0$, $O_1$ in Equation \ref{Eq.eikonal1} and $O_{2-0}$ to $O_{2-2}$ in Equation \ref{Eq.eikonal4}, we derive the losses for SDF supervision, based on our Sandwich Eikonal formulation. Considering that continuous space supervision is unfeasible, we sample the discrete points in the corresponding space ($\Omega_0$, $\Omega_1$, $\Omega_{occ}$ and $\Omega_{free}$) for supervision. 

\begin{figure}[h]
\begin{center}
\includegraphics[scale=0.9]{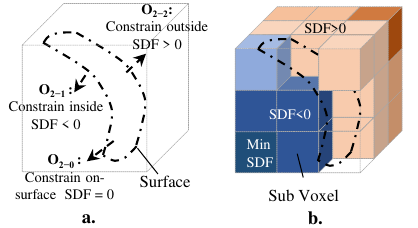}
\end{center}
\vspace{-6mm}
   \caption{\textbf{a.} SDF constraints with Sandwich Eikonal formulation in continuous form. \textbf{b.} SDF sampling in discrete form.}
\label{Fig.sample-sdf}
\vspace{-3mm}
\end{figure}
The correctness and density of the supervision sampling are of paramount importance. Since our approach is based on the surround cameras, we only sample from the camera-visible space in $\Omega_0$ (space of interest) \cite{tong2023scene}.
For $\Omega_1$ (surface points), we sample exclusively from the LiDAR points. To enhance the sampling density, temporal point cloud fusion is adopted. Specifically, $\Omega_1=\{p_{t-k}, p_{t-(k-1)}, ..., p_{t+(k-1)}, p_{t+k}\}$, where $p_{t+j}$ is the LiDAR points of the $t+j$ frame and $t$ is the current frame.
Based on occupancy GT \cite{tong2023scene}, $\Omega_{occ}$ and $\Omega_{free}$ are the sets of voxel grids labeled as ``occupied" and ``free", respectively. 

Additionally, in our objective $O_{2-1}$, the computation of the minimum SDF within the occupied grid is required. As is shown in Figure \ref{Fig.sample-sdf}, we approximate this minimum value through discrete sampling. The minimum value among the SDF values at the centers of each sub-grid is regarded as the minimum SDF value for the grid. Based on Equation \ref{Eq.eikonal1}, \ref{Eq.eikonal4} and the above sampling strategy, our SDF loss $\mathcal{L}_{sdf}$ is expressed as follows:
\vspace{-6mm}

\begin{equation}
\scalebox{0.9}{$
    \begin{split}
    \mathcal{L}_{sdf} & = \gamma_{1}\frac{1}{n_{total}} \sum_{i=1}^{n_{total}} (|\nabla _\textbf{x}\phi(\textbf{x}^i) |-1)  \\
    &  + \gamma_{2}\frac{1}{n_{1 \cup occ}} \sum_{i=1}^{n_{1 \cup occ}} (\nabla _\textbf{x}\phi(\textbf{x}^i) -n(\textbf{x}^i))  \\
    &  + \gamma_{3}\frac{1}{n_{1}}\sum_{i=1}^{n_{1}}\left|\phi(\textbf{x}^i)\right| \\
    &  + \gamma_{4}\frac{1}{n_{occ}}\sum_{i=1}^{n_{occ}}e^{\alpha \times min(\phi(\textbf{x}^i), \textbf{x}^i \in \Omega_{occ}^i)} \\
    &  +\gamma_{5} \frac{1}{n_{free}}\sum_{i=1}^{n_{free}}e^{-\alpha \times random(\phi(\textbf{x}^i), \textbf{x}^i \in \Omega_{free}^i)},
    \end{split}
    $}
\label{loss_sdf} 
\vspace{-2mm}
\end{equation}

where $\alpha$ is a hyperparameter that determines the extent of the deviation for the interior and exterior space of the object and we fix $\alpha$ to 100 in experiments. $\gamma_{1}, \gamma_{2}, \gamma_{3}, \gamma_{4}, \gamma_{5}$ are loss weights, $n(\textbf{x})$ represents the normal vector of the current point on surface. $n_{1}$, $n_{occ}$, and $n_{free}$ represent the number of points or voxel grids sampled from $\Omega_{1}$, $\Omega_{occ}$ and $\Omega_{free}$, respectively. $n_{total}$ is the total number of samples: $n_{total} = n_{1} + n_{occ} + n_{free}$. 

\vspace{-1mm}
\paragraph{Semantic Supervision} A widely used cross-entropy loss is adopted to supervise the semantic field as follows:
\vspace{-3mm}

\begin{equation}
    \scalebox{0.9}{$
        \mathcal{L}_{sem} = \frac{1}{n_{occ}}\sum_{i=1}^{n_{occ}}CE(S(\textbf{x}|\textbf{x} \in \Omega_{occ}^i), \hat{y}^{i}_{sem}),
    $}
\label{} 
\vspace{-1mm}
\end{equation}
where CE represents cross-entropy loss and $\hat{y}_{sem}^i$ is the semantic ground truth of i-th voxel grid.

\vspace{-1mm}
\paragraph{Joint Supervision} 
\label{Sec.joint}

\begin{figure}[h]
\begin{center}
\includegraphics[scale=0.45]{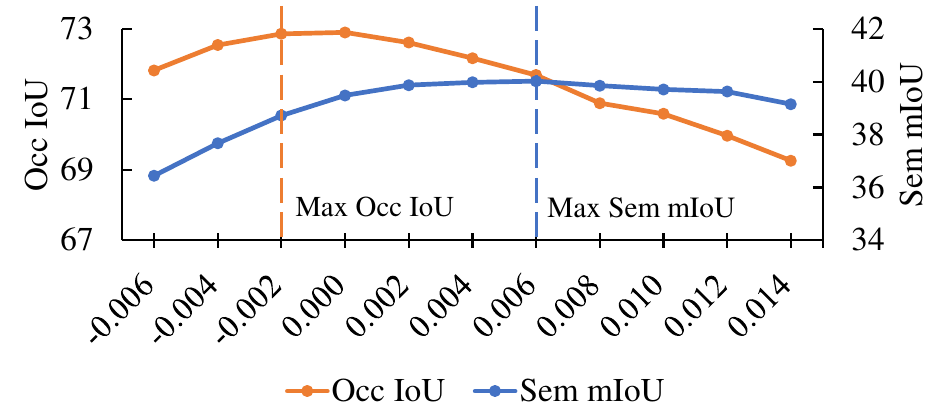}
\end{center}
\vspace{-6mm}
   \caption{Variation of occupancy IoU and semantic mIoU with SDF Threshold. Note that the peak values of these two indicators correspond to different SDF thresholds.}
\label{Fig.jointsup}
\vspace{-4mm}
\end{figure}

The above losses optimize the SDF and semantic field separately, which leads to a non-negligible issue of ambiguity between geometric and semantic optimization. To illustrate this issue, we train a model and evaluate the occupancy IoU representing geometric accuracy and the semantic mIoU representing overall semantic accuracy following the occupancy prediction metrics \cite{tong2023scene, tian2023occ3d}. For each voxel grid, if its minimum SDF is less than a certain threshold, the voxel is considered 'occupied'. As is shown in Figure \ref{Fig.jointsup}, the optimal thresholds producing the highest peak of these curves are significantly different. This indicates that greater geometric accuracy does not necessarily lead to improved overall semantic accuracy.

To mitigate this issue, we propose a joint supervision strategy. First, for the minimum SDF value $min(\phi(\textbf{x}))$ of each voxel grid, a SoftMax function is used to convert it to the free probability: $p_{free} = softmax(\beta t, \beta min(\phi(\textbf{x})))$, where t is fixed to 0.005, $\beta$ is fixed to 100. Second, we map the free probability $p_{free}$ to logits using an inv-sigmoid function: $l_{free} = log(p_{free}/(1-p_{free}))$.
Finally, $l_{free}$ is combined with semantic logits $l_{sem}$, and the joint loss $\mathcal{L}_{joint}$ is expressed as:
\vspace{-2mm}
\begin{equation}
    \mathcal{L}_{joint} = Dice(C(l_{sem},l_{free}), C(\hat{y}_{sem}, \hat{y})),
\end{equation}
where C is the concatenation operation and $\hat{y}_{sem}$ is the semantic GT of voxel grid. $Dice$ represents the Dice loss \cite{milletari2016v}. Finally, the total loss is calculated as:
\vspace{-2mm}
\begin{equation}
    \mathcal{L} = \gamma_{6} \mathcal{L}_{sdf} + \gamma_{7} \mathcal{L}_{sem} + \gamma_{8} \mathcal{L}_{joint}
\vspace{-2mm}
\end{equation}
where $\gamma_{6}, \gamma_{7}, \gamma_{8}$ are loss weights.

\subsection{Inference}
\label{Sec.inference}

Based on the output SDF and semantic field, diverse outputs for different tasks can be obtained through different sampling strategies and post-processing methods.

\vspace{-4mm}
\paragraph{Semantic Mesh Generation} 
\label{section:mesh-inference}
To generate the semantic mesh, we uniformly query the SDF value in the space of interest and use the marching cubes algorithm \cite{newman2006survey} to obtain the mesh. Subsequently, at the vertex positions of each triangular face, we query the corresponding semantic information, resulting in a mesh enriched with semantics.
\vspace{-2mm}
\paragraph{Occupancy Prediction} 
\label{section:occ-inference}
The objective of the occupancy prediction is to anticipate the semantics of the voxel grid in space. We achieve this through the joint prediction based on the predicted SDF and semantic field as follows:
\vspace{-2mm}
\begin{equation}
    \mathcal{S}_i = argmax(C(l^i_{sem}, l^i_f)), 
    \label{Equ.class} 
\end{equation}
where $\mathcal{S}_i$ is the i-th voxel grid, $l^i_{sem}$ is the logits of the i-th voxel grid,and  $l^i_f$ the logits value from SDF.

\begin{figure}[h]
\begin{center}
\includegraphics[scale=0.4]{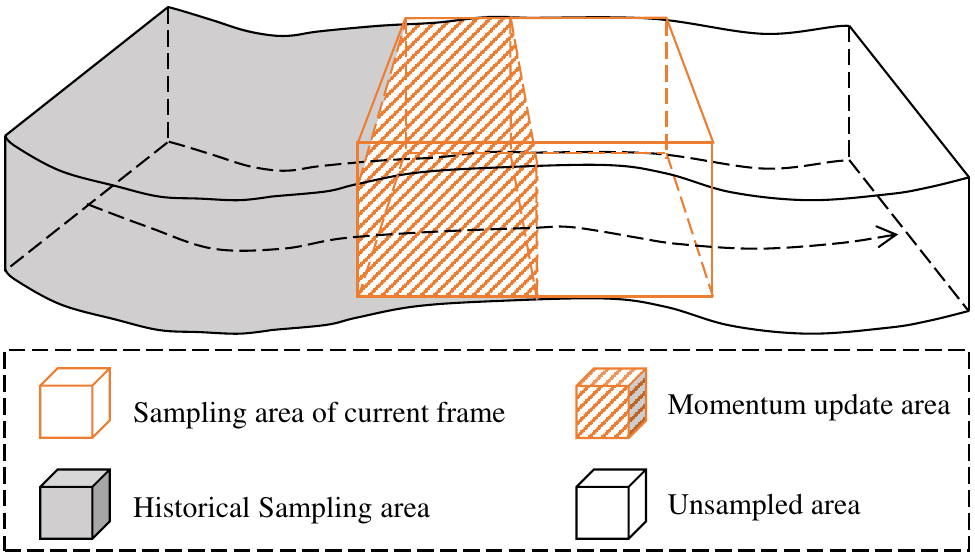}
\end{center}
\vspace{-4mm}
   \caption{The process of semantic scene reconstruction. SDF values and semantic logits are sampled long the vehicle trajectory into the world coordinate system. For regions where the historical sampling overlaps with the current frame, the historical SDF and semantic logits are updated based on a momentum update strategy.}
\label{Fig.recon-pipeline}
\vspace{-6mm}
\end{figure}

\begin{figure}[t]
\begin{center}
\includegraphics[scale=0.26]{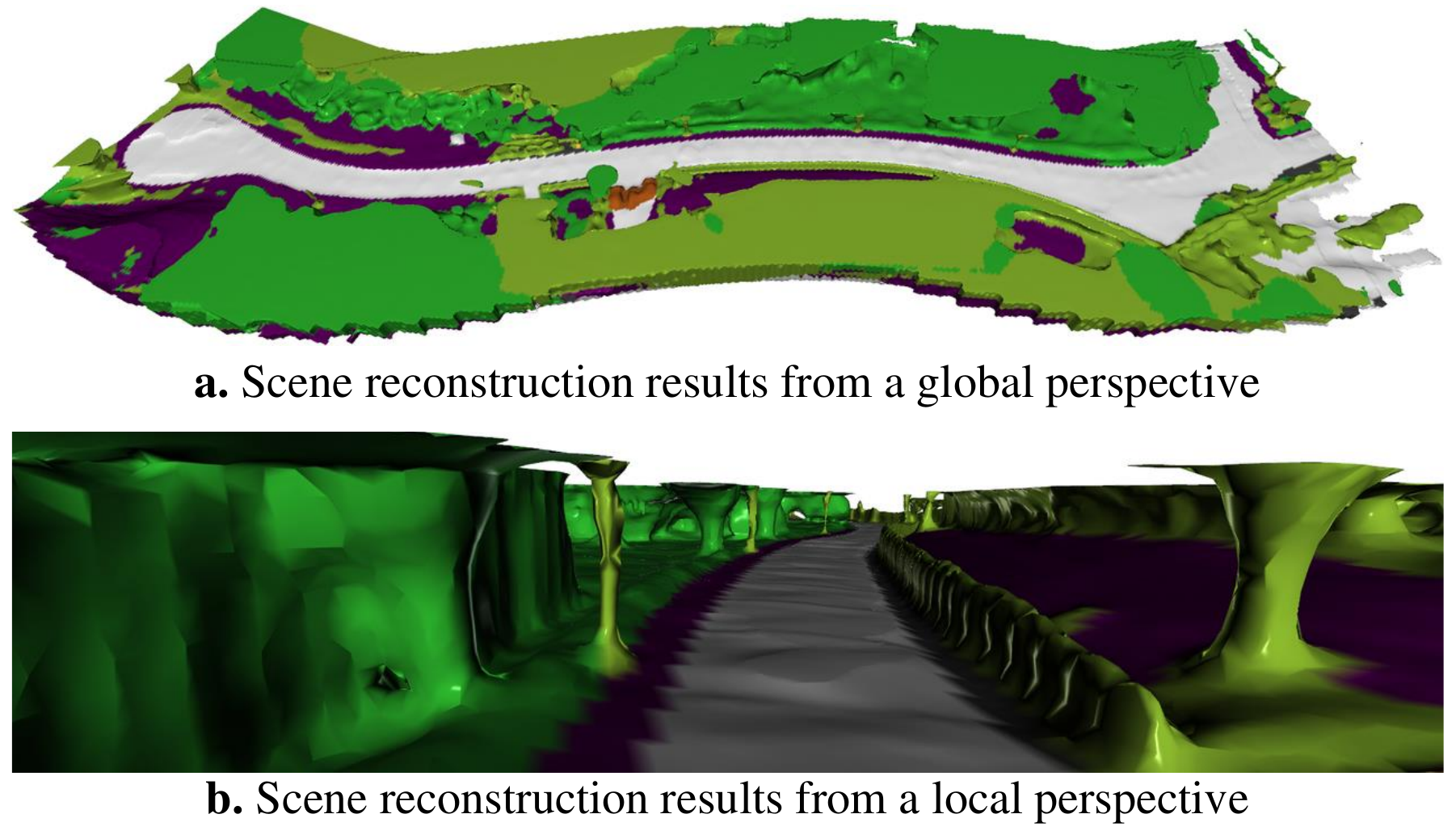}
\end{center}
\vspace{-6mm}
   \caption{Visualization of semantic scene reconstruction.}
\label{Fig.recon}
\vspace{-6mm}
\end{figure}

\vspace{-2mm}
\paragraph{Semantic Scene Reconstruction} 
Our SurroundSDF demonstrates a notable advantage by achieving semantic scene reconstruction concurrently with scene perception. As is shown in Figure \ref{Fig.recon-pipeline}, we sequentially query the semantic and SDF values for the region of interest frame by frame. The SDF values and semantic logits of static objects are projected onto the sampling points in the world coordinate system.  
For the overlapping of the historical sampling area and the current frame sampling area, we employ a momentum update strategy to refine the historical SDF values and semantic logits. Figure \ref{Fig.recon} shows the final scene reconstruction result with the marching cubes algorithm.
%utilizing the marching cubes algorithm, we obtain the final scene reconstruction results, as illustrated in Figure \ref{Fig.recon}.

\vspace{-2mm}
\section{Experiments}

\subsection{Experimental Setup}
\vspace{-2mm}

\begin{table}[h]
\small
\renewcommand{\arraystretch}{1.3}
\begin{center}
\setlength{\tabcolsep}{3pt}{
\scalebox{0.9}{
\begin{tabular}{ c | c | c | c | c}
\toprule
Version & Backbone & Image size  & Channels & Frames  \\
\cline{1-5}
\textit{Small}     &  ResNet50 \cite{he2016deep} &  $256\times704$  &  32  &  9 \\
\textit{Medium}   & ResNet101 \cite{he2016deep} & $640\times1600 $   &  128 &  6 \\
\textit{Large} &   ConvNext-B \cite{liu2022convnet}&   $640\times1600 $  &  256   &  3 \\
\bottomrule
\end{tabular}
}}
\vspace{0mm}
\caption{Experimental settings of different versions.}
\label{Tab.setting} 
\vspace{-6mm}
\end{center}
\end{table}
\noindent \textbf{Implementation details}
We follow BEVStereo \cite{li2023bevstereo} to construct our 3D encoder, where the depth net predicts 88 discrete depth values, the voxel feature resolution is $200\times200\times16$, and the temporal fusion strategy is applied. We pre-train our architecture on semantic segmentation, depth estimation, and 3D object detection tasks on the NuScenes training set. To compare with methods under different settings, we train three versions: \textit{Small}, \textit{Medium}, and \textit{Large} with different backbone, input image size, voxel feature channels, and the number of temporal frames, as is shown in Tabel \ref{Tab.setting}. For the medium and large versions, we reduce the number of temporal frames to save GPU memory. We use the Adam optimizer \cite{kingma2014adam}, set the batch size to 32, fix the learning rate to 1e-4, and train for around 32 epochs for each version. The loss weights $\gamma_{1}\sim\gamma_{8}$ are set to 1.0, 1.0, 30.0, 0.05, 0.05, 1.0, 1.0, and 1.0 respectively.

\noindent \textbf{Dataset} We evaluate our method on Occ3D-nuScenes dataset \cite{tong2023scene, tian2023occ3d}, which is an occupancy dataset with dense voxel grid label based on the nuScenes dataset \cite{caesar2020nuscenes}. It processes dynamic and static objects separately, uses multi-frame aggregation to achieve point cloud densification, and then reconstructs the dense point cloud to obtain a 3D occupancy label. It has voxel-level semantic labels of 17 classes, which contain 16 common classes and an additional general object class. Each frame covers a range of [-40m, -40m, -1m, 40m, 40m, 5.4m] with a voxel size of 0.4m, resulting in a voxel grid resolution with $200\times200\times16$. 

\noindent \textbf{Evaluation Metrics} 
To evaluate our performance on dense 3D scene perception, we follow the 3D semantic occupancy prediction task \cite{tong2023scene} and use the widely adopted mean intersection over union (mIoU). 

\begin{equation}
\scalebox{0.9}{$
    \begin{aligned}
        mIoU=\frac{1}{C}\sum_{i=1}^{C}\frac{TP_i}{TP_i+FP_i+FN_i}
    \end{aligned},
    $}
\end{equation}
where $C$ represents the number of categories (set to 17 in the experiment), and TP, FP, and FN represent true positive, false positive, and false negative respectively.

\begin{table}[h]
\small
\renewcommand{\arraystretch}{1.3}
\begin{center}
\setlength{\tabcolsep}{26pt}{
\scalebox{0.9}{$
\begin{tabular}{ c | c }
\toprule
Metrics & Formula \\
\cline{1-2}
Abs Rel & $\frac{1}{n}\sum \lvert d-d^* \rvert/d^* $    \\
Sq Rel & $\frac{1}{n}\sum \lvert d-d^* \rvert^2/d^* $   \\
RMSE & $\sqrt{\frac{1}{n}\sum \lvert d-d^* \rvert^2} $     \\
$\delta<1.25$ & $\frac{1}{n}\sum(max(\frac{d}{d^*},\frac{d^*}{d})<1.25) $ \\
\bottomrule
\end{tabular}
$}
}
\vspace{-2mm}
\caption{Depth metrics for 3D scene reconstruction}
\vspace{-5mm}
\label{Tab.depth}
\end{center}
\end{table}

To evaluate our scene reconstruction performance, we render the 2D depth image for the surround cameras based on the reconstructed scene and validate using 2D depth evaluation.
As is shown in Table \ref{Tab.depth}, we use the metrics commonly used for depth estimation, which aim to measure the difference between the predicted depth obtained by mesh rendering and GT depth obtained by LiDAR points.

\subsection{3D Semantic Occupancy Prediction}
Following the widely adopted experimental setup \cite{tong2023scene, li2023fbocc}, we conduct the semantic occupancy prediction experiment on the Occ3D-nuScenes dataset. For a fair comparison, in the inference phase, we divide the space into voxel grids with the same size as the other methods (0.4m) and predict the occupancy semantics following Section \ref{section:occ-inference}. As is shown in Tabel \ref{Tab.occ}, the mIoU score of our \textit{Small} version using ResNet50 as the backbone and without any test-time-augmentation (TTA) achieves 42.37, surpassing all the methods with the same image backbone. Additionally, our \textit{Medium} version significantly outperforms BEVFormer \cite{li2022bevformer} and VoxFormer \cite{li2023voxformer} using the ResNet101 backbone. Moreover, our \textit{Large} version surpasses UniOCC \cite{Pan_Liu_Liu_Huang_Wang_Xu_Lai_Zhang_Yang_Car} with the same ConvNext-B backbone by a 1.5 mIoU score, and even outperforms FB-OCC \cite{li2023fb, li2023fbocc}, which utilizes a larger image backbone and TTA strategy. These results demonstrated the effectiveness of SurroundSdf in the perception task.

\subsection{3D Scene Reconstruction}
\vspace{-1mm}
To evaluate the performance of scene reconstruction, we compare our method with state-of-the-art (SOTA) surround-view-based depth estimation methods \cite{wei2023surrounddepth, schmied2023r3d3} and occupancy prediction methods \cite{wei2023surroundocc, li2023fbocc}. To acquire the surround depth, we utilize the semantic mesh generation steps described in Section \ref{section:mesh-inference} to generate the mesh and then render the depth maps for each camera based on the camera parameters. To fairly compare with the occupancy-based methods, we follow a similar process to render the depth maps with predicted occupied voxel grids, and all methods use RestNet50 as the image backbone. As is shown in Table \ref{Tab.Recon}, our method exhibits significant advantages over occupancy-based approaches across all metrics. Moreover, on the primary metric of absolute relative error, we achieve SOTA results, surpassing SurroundDepth \cite{wei2023surrounddepth} and R3D3 \cite{schmied2023r3d3}, which are the SOTA methods for surround-view depth estimation.

% Comparsion on the Occ3D-nuScenes val set.
\begin{table}[h]
\small
\renewcommand{\arraystretch}{1.3}
\begin{center}
% \label{Ablation}
\scalebox{0.95}{$
\begin{tabular}{ l | c | c | c}
\toprule
\multicolumn{1}{l|}{\bf{Methods}} &
\multicolumn{1}{c|}{\bf{Backbone}} & \multicolumn{1}{c|}{\bf{Image Size}}& \multicolumn{1}{c}{\bf{mIoU}} \\
\cline{1-4}

SurroundOcc \cite{wei2023surroundocc} & ResNet50  &  $256\times704$ & 36.32 \\

BEVStereo \cite{li2023bevstereo} & ResNet50  &  $384\times704$ & 39.1 \\

UniOCC \cite{Pan_Liu_Liu_Huang_Wang_Xu_Lai_Zhang_Yang_Car} & ResNet50 &  $256\times704$ &  39.7 \\ 

MiLO \cite{Vu_Kim_Kim_Jung_Jeong}  & ResNet50  &  $256\times704$ & 40.49 \\

FB-OCC \cite{li2023fbocc} & ResNet50  & $256\times704$  & 40.69 \\

FB-OCC\textsuperscript{*}  \cite{li2023fbocc} & ResNet50  & $256\times704$  & 42.06 \\

Ours & ResNet50  & $256\times704$  & \bf{42.37}\\

\hline

BEVFormer \cite{li2022bevformer} &  ResNet101 & $900\times1600$  & 40.6 \\

VoxFormer \cite{li2023voxformer} &  ResNet101 & $900\times1600$  & 40.6 \\

Ours & ResNet101  &  $640\times1600$ & \bf{46.0} \\

\hline

SurroundOcc \cite{wei2023surroundocc} &  InterImage-B & $640\times1600$  & 40.7 \\

BEVFormer \cite{li2022bevformer} & InterImage-XL  &  $640\times1600$ & 43.3 \\

BEVDet \cite{huang2021bevdet} & Swin-B  &  $640\times1600$ & 43.1 \\

UniOCC \cite{Pan_Liu_Liu_Huang_Wang_Xu_Lai_Zhang_Yang_Car} & ConvNext-B & $640\times1600$  &  51.5 \\ 

FB-OCC\textsuperscript{*} \cite{li2023fbocc} & InternImage-H & $960\times1769$   & 52.79 \\

Ours & ConvNext-B  &  $640\times1600$ & \bf{53.01} \\

\bottomrule
\end{tabular}
$}

\caption{Comparsion on the Occ3D-nuScenes val set. The superscript * denotes using test time augmentation.}
\vspace{-6mm}
\label{Tab.occ}
\end{center}

\end{table}

\begin{table}[h]
\small
\begin{center}
\setlength{\tabcolsep}{3pt}{
\begin{tabular}{ c | c | c | c | c}
\toprule
Methods & Abs Rel $\downarrow$ & Sq Rel $\downarrow$ & RMSE $\downarrow$ & $\delta<1.25\uparrow$  \\
\cline{1-5}
SurroundOcc   &    0.274 &  2.072   &  5.327   &  0.482 \\
FB-OCC        &   0.342  &  2.047   &   5.970  &  0.290 \\
SurroundDepth &   0.224  &   2.102  &  4.573   &  \textbf{0.751} \\
R3D3          &   0.259  &  2.328   &  5.577   &  0.583 \\
Ours          &   \textbf{0.174}  &   \textbf{1.097}  &  \textbf{4.402}   &  0.747 \\
\bottomrule
\end{tabular}}
\vspace{-2mm}
\caption{3D scene reconstruction performance. AbsRel is the main metric. ``$\downarrow$" means less is better.}
\vspace{-4mm}
\label{Tab.Recon}
\vspace{-5mm}
\end{center}
\end{table}

\subsection{Ablation Study}
In this section, we study the proposed strategies and the impact of GT quality.

\begin{table*}[ht]
\small
\renewcommand{\arraystretch}{1.3}
\begin{center}
% \label{Ablation}
\setlength{\tabcolsep}{2pt}{
\resizebox{0.93\linewidth}{!}{
\begin{tabular}{ c | c  c  c | c  c  c  c  c  c  c  c  c  c  c  c  c  c  c  c  c | c | c }
\toprule
\bf{Model} & \rotatebox{90}{3D Representation}  & \rotatebox{90}{SDF Supervision} & \rotatebox{90}{Joint Sup} & \rotatebox{90}{others} & \rotatebox{90}{barrier}  & \rotatebox{90}{bicycle}  & \rotatebox{90}{bus}  & \rotatebox{90}{car}  & \rotatebox{90}{vehicle}  & \rotatebox{90}{motorcycle}  & \rotatebox{90}{pedestrian}  & \rotatebox{90}{traffic cone}  & \rotatebox{90}{trailer}  & \rotatebox{90}{truck}  & \rotatebox{90}{driveable surface}  & \rotatebox{90}{other flat}  & \rotatebox{90}{sidewalk}  & \rotatebox{90}{terrain}  &  \rotatebox{90}{manmade} & \rotatebox{90}{vegetation}  & \bf{mIoU} & \bf{Abs-Rel}\\
\cline{1-23}

A & OCC & $\rule[0.5ex]{3mm}{0.2pt}$ & $\rule[0.5ex]{3mm}{0.2pt}$ &  8.2 &  47.0  & 7.6  &  43.6 &  51.4 &  25.6 & 18.4  & 21.7  & 22.4  & 32.2 & 38.7 & 83.6  & 43.0  & 55.3 & 57.7  & 50.6  & 44.6  & 38.33 & 0.342   \\ 
B & SDF & SIREN & $\rule[0.5ex]{3mm}{0.2pt}$ &  8.0 &  41.4  & 18.6  &  25.9 &  41.0 &  19.5 & 19.0  & 15.6  & 23.0  & 19.8 & 27.8  & 72.3  & 38.6  & 48.7 & 46.5  & 36.6  & 33.6  & 31.51   & 0.188\\ 
C & SDF & LODE & $\rule[0.5ex]{3mm}{0.2pt}$ &  8.7 &  46.0  & 18.6  &  38.5 &  47.2 &  25.3 & 22.2  & 20.4  & 26.3  & 23.7 & 34.5  & 81.2  & 44.8  & 54.0 & 53.8  & 40.3  & 37.8  & 36.66  & 0.192  \\ 
D & SDF & Sandwich & $\rule[0.5ex]{3mm}{0.2pt}$ &  12.3 &  47.9  & 25.4  &  43.0 &  51.7 &  27.4 & 24.9  & 24.9 & 27.4  & 34.0  & 39.4  & 83.6  & 42.9  & 54.4 & 58.1  & 48.3  & 43.2  & 40.52 &  \bf{0.171} \\ 
E & SDF & Sandwich & \checkmark &  13.9 &  49.7  & 27.8  &  44.6 &  53.0 &  30.0 & 29.0  & 28.3  & 31.1  & 35.8 & 41.2  & 83.6  & 44.6  & 55.3 & 58.9  & 49.6  & 43.8  & \bf{42.37} & 0.174  \\ 

\bottomrule

\end{tabular}
}
}
\vspace{-3mm}
\caption{Results of the ablation experiments. ``3D Representation" indicates whether the scene is described by occupancy voxels or SDF. ``SDF Supervision" indicates the supervised paradigm. ``Joint Sup" indicates whether the joint supervision strategy in Section \ref{Sec.joint} is used.}
\vspace{-7mm}
\label{Tab.ablation}

\end{center}
\end{table*}

\label{Sec.ablation}
\noindent \textbf{Ablation Study on Proposed Strategies} We conduct ablation experiments based on the settings of the \textit{Small} version in Tabel \ref{Tab.setting} and the results are shown in Tabel \ref{Tab.ablation}. These studies aim to analyze the effectiveness of different 3D representations and supervisions. Specifically, we first construct a baseline model (model A) that is supervised by the occupancy GT with the cross-entropy loss and output the occupancy results. Then, to investigate the 3D perception capability of implicit SDF and semantic field, we build the subsequent models based on the architecture in Section \ref{Sec.arch} and output the SDF and semantic field. To analyze the proposed weak supervision paradigm with previous methods, we reproduce the supervision methods of SIREN (model B) and LODE (model C). In model D, we employ our Sandwich Eikonal for SDF supervision. Lastly, model E incorporates the joint supervision strategy in section \ref{Sec.joint}. 

Comparing the results of Model B, and C with A, we observe a significant advantage in the AbsRel metric for both models B and C. This highlights the robust reconstruction capability of SDF-based 3D representation. However, we also observe a significant decline in the semantic mIoU metric for model B, attributed to the sparse supervision with LiDAR points. LODE mitigates this issue by introducing occupancy supervision, but its mIoU score is still lower than the baseline. Using the proposed Sandwich Eikonal supervision paradigm, the results of model D not only surpass the baseline in semantic mIoU but also outperform models B and C in AbsRel. This indicates the effectiveness of our supervision approach, which successfully integrates the supervision from LiDAR points and occupancy GT. Furthermore, by incorporating of the proposed joint supervision strategy, model E exhibits a 1.85 improvement in semantic mIoU, accompanied by a slight regression of 0.003 in the AbsRel metric. For some long-tail categories, such as bicycle, motorcycle, and traffic cone, our joint supervision strategy yields a noticeable improvement.
\begin{figure*}[ht]
\begin{center}
\includegraphics[scale=0.47]{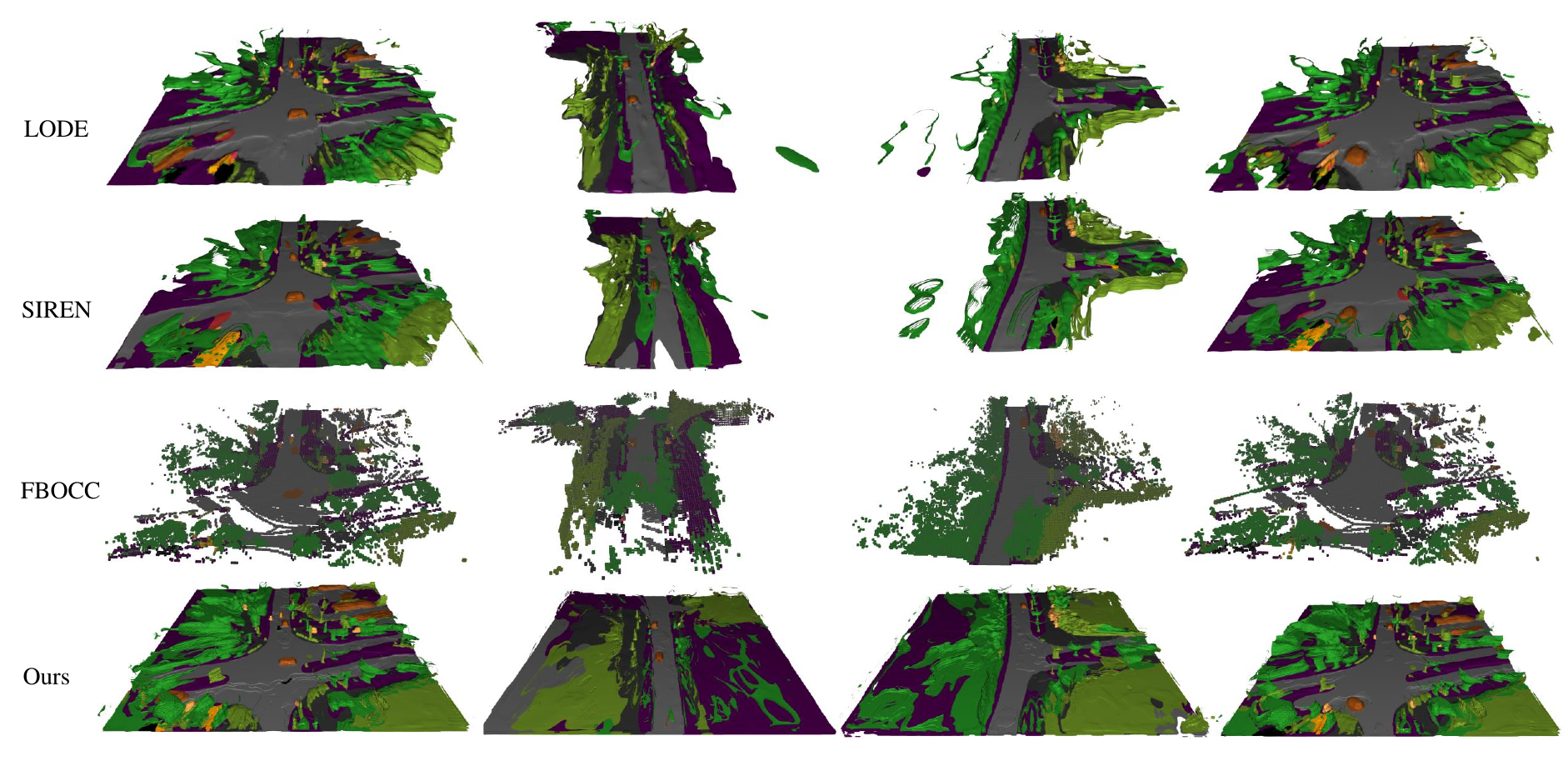}
\end{center}
\vspace{-7mm}
   \caption{Comparison of the 3D scene perspective results with SOTA methods.}
\vspace{-4mm}
\label{Fig.vis}
\end{figure*}

\begin{table}[h]
\centering
\setlength{\tabcolsep}{2.5pt}{
\scalebox{0.86}{
\begin{tabular}{l|cc|cc|cc}
\hline
OCC GT size & \multicolumn{2}{c|}{0.4m} &\multicolumn{2}{c|}{0.8m} & \multicolumn{2}{c}{1.6m} \\
\hline
Target & mIoU & Abs-Rel & mIoU & Abs-Rel & mIoU & Abs-Rel \\
\hline
 FBOCC & 40.69 & 0.342 & 36.01 & 0.370 &26.96  & 0.445 \\
 SurroundSDF& 42.37 & 0.174 & 40.74 & 0.176 & 40.16 & 0.181 \\
\hline
\end{tabular}}}
\vspace{-2mm}
\caption{Results of the experiments on the resolution of OCC GT.}
\vspace{-1mm}
\label{tab:occ_gt}
\end{table}

\begin{table}[h]
\small
\begin{center}
\setlength{\tabcolsep}{8pt}{
\begin{tabular}{ c  c  c  c  c}
\toprule
Sampling rate & 1 & 0.5 & 0.2 & 0.1\\
\cline{1-5}
mIoU   &    42.37 &  42.37   &  42.16   &  41.07 \\
Abs-Rel  &   0.174  &  0.170   &   0.186  &  0.186 \\ 
\hline
\end{tabular}}
\vspace{-2mm}
\caption{Ablation study on the density of LiDAR points GT.}
\vspace{-8mm}
\label{Tab:pcd}

\end{center}
\end{table}

\noindent \textbf{Ablation Study on Ground Truth Quality} 
To further validate the roles of OCC GT and LiDAR GT, we conducted ablation experiments on the resolution of OCC GT and the density of LiDAR points respectively. As is shown in Table \ref{tab:occ_gt}, the impact of OCC GT resolution variations in our method's OCC GT is mitigated with LiDAR points supervision. When the voxel grid resolution increased from 0.4m to 1.6m, the mIoU only decreased by 5.2\%, and Abs-Rel increased by a mere 4.0\%. In comparison, the SOTA OCC prediction method FBOCC experienced a substantial mIoU decrease of 33.7\%, coupled with a 30.1\% increase in Abs-Rel. 
As shown in Table \ref{Tab:pcd}, utilizing only 10\% of the point cloud resulted in a marginal decrease of 3.1\% in mIoU and a modest increase of 6.9\% in Abs-Rel. These experiments further substantiate the robustness and efficacy of the proposed Sandwich Eikonal supervision paradigm.

\subsection{Comparison of Visual Results}
\vspace{-2mm}
We visualize our 3D scene perception results and compare them with the reproduced SIREN \cite{sitzmann2020implicit} and LODE \cite{LiZSZYZZ23}  (models B and C in Section \ref{Sec.ablation}), and the SOTA occupancy prediction method FBOCC \cite{li2023fb, li2023fbocc}, as illustrated in Figure \ref{Fig.vis}. Our SurroundSDF demonstrates significant advantages in the continuity and accuracy of the geometric structure.

\vspace{-1mm}
\section{Conclusion}
\vspace{-1mm}

In this work, we propose SurroundSDF, a novel vision-centric 3D scene understanding framework. We introduce SDF to address the continuity and accuracy of perception from surround cameras. Moreover, in the absence of SDF GT, we propose Sandwich Eikonal formulation, a novel SDF supervision paradigm, which emphasizes imposing appropriate constraints on both sides of the surface to enhance geometric accuracy. Furthermore, to alleviate the inconsistency between geometric optimization objectives and semantic optimization objectives, we introduce a joint supervision strategy. Based on the generated SDF and semantic field, different 3D representations can be obtained, including the scene mesh, occupancy voxel grids, and semantic reconstruction mesh of the whole scene. Comprehensive experiments validate the performance of our method.

{

    \small
    \bibliographystyle{ieeenat_fullname}
    \bibliography{egbib}
}
% \input{X_suppl}
% WARNING: do not forget to delete the supplementary pages from your submission 
% \input{sec/X_suppl}

\end{document}